\pdfoutput=1

\documentclass[11pt]{article}

\usepackage{EMNLP2023}

\usepackage{times}
\usepackage{latexsym}
\usepackage{graphicx}
\usepackage{linguex}
\usepackage{amsmath,amssymb}

\usepackage{booktabs, multirow} 
\usepackage{soul}

\usepackage[T1]{fontenc}

\usepackage[utf8]{inputenc}

\usepackage{titlesec}
\titlespacing*{\section}{0pt}{0.3\baselineskip}{0.3\baselineskip}
\usepackage{microtype}

\usepackage{inconsolata}

\title{Sociolinguistically Informed Interpretability: \\
A Case Study on Hinglish Emotion Classification}

\author{Kushal Tatariya$^{1}$ \ \ Heather Lent$^{2}$ \ \ Johannes Bjerva$^{2}$  \ \ Miryam de Lhoneux$^{1}$ \\
         $^1$ Department of Computer Science, KU Leuven, Belgium  \\
          $^2$ Department of Computer Science, Aalborg University, Denmark \\
           \texttt{\{kushaljayesh.tatariya, miryam.delhoneux\}@kuleuven.be}\\ 
           \texttt{\{hcle, jbjerva\}@cs.aau.dk} \\
           }

\begin{document}
\maketitle
\begin{abstract}

Emotion classification is a challenging task in NLP due to the inherent idiosyncratic and subjective nature of linguistic expression,
especially
with code-mixed data. 
Pre-trained language models (PLMs) have achieved high performance for many tasks and languages, but it remains to be seen whether these models learn and are robust to the differences in emotional expression across languages. 
Sociolinguistic studies have shown that Hinglish speakers switch to Hindi when expressing negative emotions and to English when expressing positive emotions. To understand if language models can learn these associations, 
we study the effect of language on emotion prediction across 3 PLMs on a Hinglish emotion classification dataset. Using LIME \cite{DBLP:journals/corr/RibeiroSG16} and token level language ID, we find that models do learn these associations between language choice and emotional expression. Moreover, having code-mixed data present in the pre-training can augment that learning when task-specific data is scarce. We also conclude from the misclassifications that the models may overgeneralise this heuristic to other infrequent examples where this sociolinguistic phenomenon does not apply.

\textit{\textbf{Disclaimer:} This paper contains some examples of language use that readers may find offensive.}

\end{abstract}



\begin{figure}
\captionsetup{font=small}
    \centering
    \includegraphics[scale=0.22]{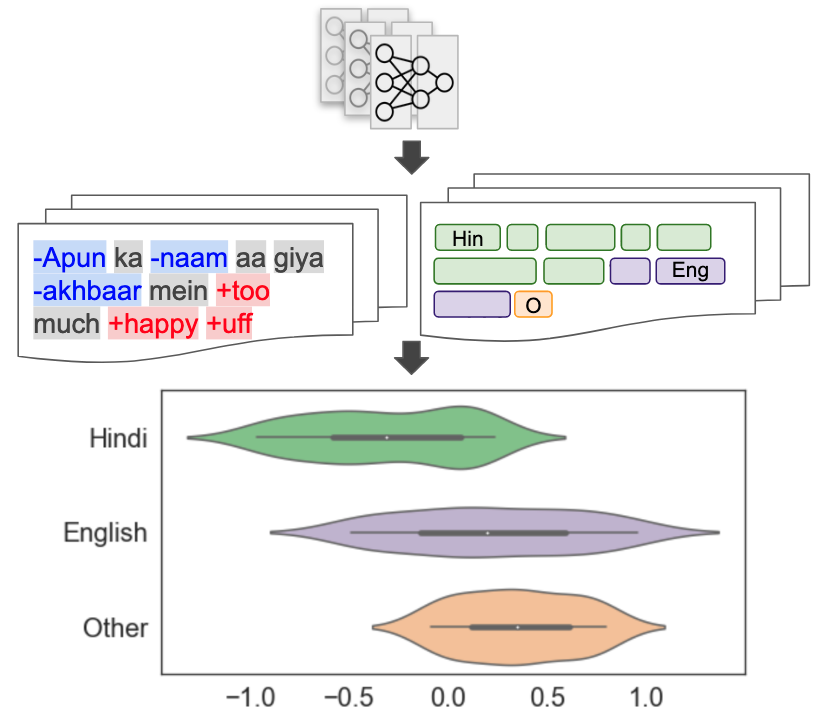}
    \caption{Our workflow. We train 3 emotion classification models, then obtain LIME scores for each token (positive scores in red, negative scores in blue, and zero scores in grey). These same samples are then tagged with token-level language ID, which enables us to examine LIME distributional differences by language.}
    \label{fig:overview}
\end{figure}

\section{Introduction}
An open-ended goal of the NLP community is to develop language technologies robust to the vast and various idiosyncrasies of authentic human communication. Understanding emotion requires knowledge of the subtleties of linguistic expression and inherent human subjectivity, making emotion classification a challenging task. It is further complicated when working with code-mixed utterances. Every language participating in code-mixed communication comes with its own cultural and linguistic baggage that oversees the verbalization of emotion \cite{KACHRU+1978+27+46, hershcovich-etal-2022-challenges}. The adoption of pre-trained language models (PLMs) has improved performance across the board for this task, but the PLMs still remain black boxes. While research in interpretability aims to address this shortcoming, most analyses remain centered around English \cite{ruder-etal-2022-square}. 
In this work, we aim to make explicit what associations are learned when PLMs are trained on code-mixed data, and
whether established differences in linguistic expression across languages indeed influence model prediction.

We approach this interpretability problem through the lens of sociolinguistics. In particular, we focus on Hindi-English (Hinglish) code-mixing, prevalent in India and in the Indian diaspora \cite{doi:10.1080/00020184.2015.1045721}. 
In a study on Hindi-English bilinguals on Twitter, \citet{rudra-etal-2016-understanding} observed that English was the language of choice for expression of a positive emotion and Hindi was more used for negative emotion. Moreover, Hindi was the preferred language for swearing online, a finding also echoed by \citet{Agarwal17}. \citet{rudra-etal-2016-understanding} explain the reason behind this to be the fact that bilinguals prefer to express strong emotions \cite{alma9977251920101488} and swear \cite{swearing_article} in L1, which happens to be Hindi for most Hinglish speakers. Conversely, \citet{rudra-etal-2016-understanding} speculate that since English is the language of aspiration in India, it becomes the preferred language for positive emotion.

In this context, we formulate our main questions as: 
(\textbf{RQ1}) Are PLMs likely to associate different emotions with different languages? 
(\textbf{RQ2}) Are English tokens more likely to influence a model to predict a positive emotion?
(\textbf{RQ3}) Are Hindi tokens more likely to influence a model to predict a negative emotion, and if so, what is the role of Hindi swear words? 
To this end, we fine-tune 3 different PLMs on a Hinglish emotion classification dataset and leverage LIME and token-level language identification for an interpretability analysis.

\section{Related Work}
\paragraph{Code-Mixing}
Previous works in emotion classification and sentiment analysis have demonstrated that processing code-mixed text is more difficult than monolingual text \cite{DBLP:journals/corr/abs-1904-00784, zaharia-etal-2020-upb, Yulianti2021}. 
This is in part due to the complexities of processing emotion from two different languages with varying socio-cultural and grammatical structures at play
\cite{9345887, SASIDHAR20201346, emotion-urdu-english}. In this context, \citet{dogruoz-etal-2021-survey} published a survey on the linguistic and social perspectives on code-mixing for language technologies. They emphasized the importance of incorporating the social context of a code-mixed language pair into systems processing code-mixed text.

\paragraph{Interpretability with LIME}
LIME (Local Interpretable Model-Agnostic Explanations) \cite{DBLP:journals/corr/RibeiroSG16} is a popular tool for interpretability that is model agnostic and employable for classification tasks. 
It learns a linear classifier locally around a model's prediction, leveraging token weights (also learned by the linear classifier) to assign a "LIME score" between 1 and -1 to each token. 
A positive score indicates that the token influenced the model towards the predicted label, and a negative score indicates that the token influenced the model to \textit{not} predict that label. 
We leverage LIME due to its availability and easy-to-use implementations, for instance in the Language Interpretability Tool \cite[LIT;][]{tenney-etal-2020-language}, which we used for this work. 
Previous work has also indicated the accuracy of its approximation of the models and its ability to provide human-friendly explanations \cite{10.1145/3546577, 10.1007/978-3-031-18344-7_42}.

\section{Methodology}\label{methods}

\paragraph{Dataset}
We utilize a dataset for sentiment analysis of code-mixed tweets by \citet{patwa-etal-2020-semeval}, later annotated with emotion labels by \citet{GHOSH2023110182}. 
Each example is annotated with the six basic Ekman emotions \cite{ekman1999basic} - \textit{joy}, \textit{sadness}, \textit{fear}, \textit{surprise}, \textit{disgust} and \textit{anger}. 
When an example does not fit any of these emotions, or expresses no emotion, it is labelled as \textit{others}. 
This dataset contains 14,000 examples in the train set, 3,000 in the validation, and 3,000 in the test set. 
For this work, we randomly sample 1,000 examples from the validation set to enable manual verification of the automatic token tagging described below, maintaining the default distribution across labels (see Appendix \ref{appendix:labels}).

\paragraph{Models}
We fine-tune 3 different PLMs for the task of emotion classification with the Hinglish training data: 
\begin{itemize}
    \item
    \textbf{XLMR} \cite{conneau-etal-2020-unsupervised}, pre-trained on Common Crawl, spanning 100 languages, including English and Hindi, both in the Devanagari script and additional romanized Hindi.
    \item
    \textbf{IndicBERT v2} \cite{doddapaneni2022indicxtreme}, pre-trained on data from 24 Indic languages, including Hindi and English that is local to the Indian subcontinent. For Hindi, this model has only seen the Devanagari script, and no romanized Hindi.
    \item
    \textbf{HingRoBERTa} \cite{nayak-joshi-2022-l3cube}, an XLM-R model that has been further pre-trained on romanized, code-mixed Hindi-English. Thus, in addition to having seen romanized Hindi, this model is specifically intended for code-mixed text.
\end{itemize}
The full details on model training and performance are given in Appendix~\ref{appendix:model}.

\paragraph{Token Tagging}
For each of the 1,000 samples (20,835 tokens in total) drawn from the validation set, we first obtain LIME scores for each token using LIT \cite{tenney-etal-2020-language}. We then run the samples through CodeSwitch \cite{github}, a Hinglish language identification tool which tags each token as Hindi, English, or Other, to get the language ID tags.\footnote{Besides the Hindi and English labels, CodeSwitch also tags tokens as "Named-Entity", "Foreign words", and "Other" for punctuation, emojis, and other non-textual tokens. For this work, we combine these 3 additional tags into one category.} 

\begin{figure}
\captionsetup{font=small}
   \centering
    \includegraphics[width = \linewidth]{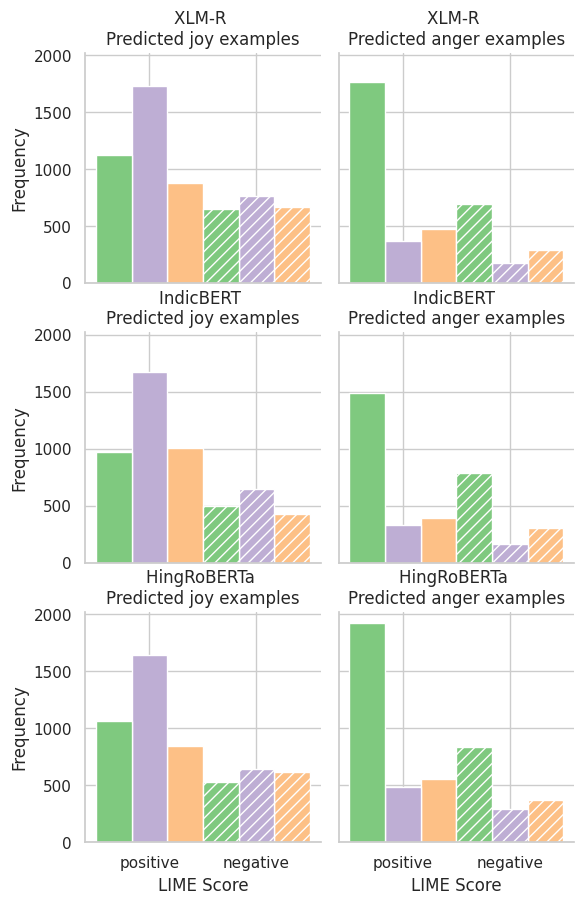}
    \caption{Frequencies of {\color{teal}\textit{Hindi}} (green), {\color{violet}\textit{English}} (purple) and {\color{orange}\textit{Other}} (orange) tokens to be assigned a positive (solid) or a negative (striped) LIME score for examples predicted as \textit{joy} and \textit{anger}, for all models.}
    \label{fig:chisquare-label}
\end{figure}

\interfootnotelinepenalty=10000

\section{Results and Analysis}
First, to answer whether the models learn to meaningfully distinguish between languages for emotion prediction (\textbf{RQ1}), we examine the distributions of LIME scores across each language ID tag (\textit{English, Hindi, Other}). 
Concretely, we inspect the frequency with which tokens received a positive or a negative LIME score in our sample, for each language. We then conduct a $\chi^2$ test of independence to determine whether these two variables have some dependency. Table \ref{tab:chisquare} shows the $p$-values for the entire sample. For all models, we find this dependence to be statistically significant ($p < 0.05$), indicating that there is some influence of language over the LIME scores. We also confirm this with a 1-Way ANOVA test, which can be found in Appendix~\ref{appendix:anova}, along with our entire statistical analysis.

Next, we examine this dependency on a more granular level to determine whether the presence of 
English tokens influence the Hinglish emotion classification models to predict more positive emotions (\textbf{RQ2}), and whether Hindi tokens influence them to predict negative emotions (\textbf{RQ3}). 
We observe the distribution of language ID across LIME scores for examples that the models predicted as \textit{joy}, \textit{anger}, and \textit{sadness}. These labels were selected in particular as they have the most examples in the dataset (after \textit{others}), and provide the positive (\textit{joy}) and negative (\textit{anger} and \textit{sadness}) polarity discussed in the sociolinguistics literature.

\begin{table}[!tp]\centering
\captionsetup{font=small}
\scriptsize
\begin{tabular}{lrrrrr}\toprule
&\multicolumn{4}{c}{\textbf{$p$-values}} \\\cmidrule{2-5}
\textbf{Model} &\textbf{Entire Sample} &\textbf{Joy} &\textbf{Anger} &\textbf{Sadness} \\\midrule
\textbf{XLM-R} &7.06e-12 &1.44e-15 &6.18e-7 &1.78e-3 \\
\textbf{IndicBERT} &1.22e-22 &3.28e-4 &1.69e-5 &3.30e-1 \\
\textbf{HingRoBERTa} &3.30e-7 &4.00e-18 &1.71e-8 &2.18e-5 \\
\bottomrule
\end{tabular}
\caption{We test the null hypothesis that language ID tags and LIME scores are independent of each other using $\chi^2$. This table contains the $p$-values for tests done on the entire sample, and also on examples predicted as \textit{joy}, \textit{anger}, and \textit{sadness}.}\label{tab:chisquare}
\end{table}

\paragraph{(RQ2) Do English tokens influence models to predict positive emotions?}
 
Figure \ref{fig:chisquare-label} shows which languages tend to have more positive and more negative LIME scores. 
As observed for \textit{joy}, English tokens have the highest frequency with positive LIME scores. 
Table \ref{tab:chisquare} shows that there is a significant dependency between language ID and LIME score for all models. 
Thus, English tokens influence the model significantly more than Hindi and Others when predicting \textit{joy}.

\paragraph{(RQ3) Do Hindi tokens influence models to predict negative emotions?}

When predicting \textit{anger}, Table \ref{tab:chisquare} again shows that there is dependency between language ID and LIME score for all models. From Figure \ref{fig:chisquare-label}, we can see that Hindi tokens influence the model significantly towards predicting \textit{anger}. When predicting \textit{sadness}, however, we only observe significance with the XLM-R and HingRoBERTa models, but not with IndicBERT. Moreover, for XLM-R, the $p$-value is not much lower than the threshold. Thus, we cannot make strong conclusions for this label.

\begin{table}\centering
\captionsetup{font=small}
\small
\begin{tabular}{lrrr}\toprule
\textbf{Token} &\textbf{Lang\_ID} &\textbf{Swear Word?\footnotemark} \\\midrule
Fuck &eng &Yes \\
Chutiye &hin &Yes \\
Fakeionist &eng &No \\
Bsdk &hin &Yes \\
Sadly &eng &No \\
Bakwas &hin &No \\
Kutta &hin &Yes \\
Gaddar &hin &No \\
Shame &eng &No \\
Sala &hin &Yes \\
\bottomrule
\end{tabular}
\caption{Top 10 tokens with the highest LIME scores when predicting negative emotions, (\textit{anger, sadness, disgust }and\textit{fear}) for all models. They have been mapped to a canonical form and are in descending order of LIME score.}\label{tab:swear-words}
\end{table}

\footnotetext{As decided by a native speaker, and also compared with the lexicon lists of Hindi and English swear words used by \citet{Agarwal17}.}

\paragraph{Swear Words} 
Previous works demonstrate that Hinglish speakers prefer to swear in Hindi over English, in a code-mixed setting \cite{rudra-etal-2016-understanding, Agarwal17}.
To check whether this finding is similarly echoed by our fine-tuned models, we examine the top 10 tokens with the highest LIME scores when predicting a negative emotion (\textit{anger}, \textit{sadness}, \textit{disgust}, \textit{fear}), across all models (see Table~\ref{tab:swear-words}). 
While the first among these is an English swear word 
(owing to it being the most used swear word by Hinglish speakers online \cite{Agarwal17}) 
there are 4 Hindi swear words in this list of tokens. 
As such, we can see that 
the models not only learn the negative connotation of the Hindi swear words, 
but also that these Hindi swear words are the \textit{most} negative of all other tokens, regardless of language, thus confirming observations from the sociolinguistics literature.

\section{Discussion}

\begin{figure}
\captionsetup{font=small}
    \centering
    \includegraphics[width=\linewidth]{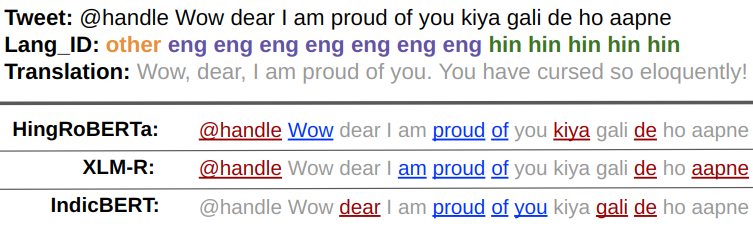}
    \caption{An example from the dataset labelled as \textit{joy}, with the translation and language ID tags. The 3 tokens with the highest LIME scores are marked in blue, and the 3 tokens with the lowest scores are marked in red.}
    \label{fig:joy-example}
\end{figure}

From the section above, it can be concluded that the models are able to distinguish patterns of speaker preference detailed by \citet{rudra-etal-2016-understanding} when predicting emotion for code-mixed data. English tokens influence the models more towards predicting a positive emotion, and Hindi tokens influence the models more towards predicting a negative emotion. An example of this is provided in Figure~\ref{fig:joy-example}, where all the models exhibit a strong degree of influence from the English tokens in their prediction of the \textit{joy} label. 
At the same time, most of the tokens assigned a negative LIME score come from Hindi.



For \textit{sadness}, we surmise that a shortage of training data is responsible for models' failure to learn meaningful differences across the languages. 
About 10\% of the entire dataset consists of examples labelled \textit{sadness}. In contrast, \textit{joy} is 30\% and \textit{anger} is about 20\% (see Appendix \ref{appendix:labels}). Even with less data, however, we still observe a dependency between language and LIME score with HingRoBERTa. 
It is the only model we examine with code-mixed data present in the pre-training. 
Thus, when there is less data for a model to learn these associations, it can help to have code-mixed data in the pre-training. 

\subsection{Do PLMs overgeneralize these learnt associations?}

\begin{figure}
    \captionsetup{font=footnotesize}
    \centering
    \includegraphics[width=\columnwidth]{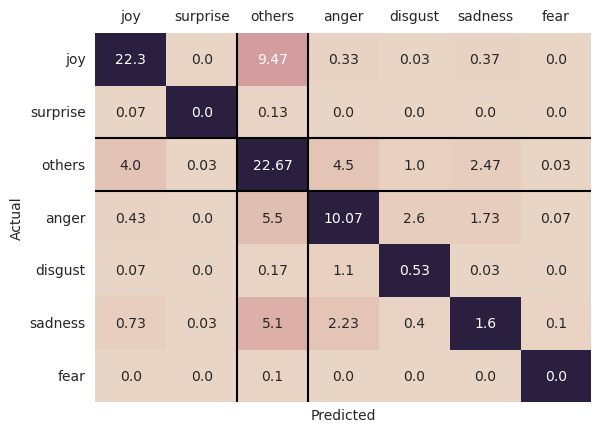}
    \caption{Confusion matrix containing the percentage of correctly and incorrectly classified examples for each label combination. The blue cells represent correct classifications, and the pink cells represent incorrect classifications.}
    \label{fig:confusion_matrix}
\end{figure}

\citet{mccoy-etal-2019-right} found that language models can adapt to heuristics that are valid for frequent cases and fail on the less frequent ones.
In a similar vein, we investigate whether these sociolinguistic associations learnt by the models overgeneralise to the less frequent examples where this phenomenon is not seen.
We examine instances where the models have misclassified examples labelled as \textit{joy} and \textit{anger}, highlighted in Figure \ref{fig:confusion_matrix}. 

For both \textit{joy} and \textit{anger}, the models generally either predict another label of the same emotional polarity (for example, \textit{disgust} instead of \textit{anger}), or they predict them as \textit{others}. The dataset is highly imbalanced, and thus we can say that although the models can discern the polarity difference between positive and negative emotion labels (as seen in Figure \ref{fig:confusion_matrix} where the values in the lower left and upper right quadrants are low), they struggle with granular distinctions between them. 

\begin{figure}
    \captionsetup{font=footnotesize}
    \centering
    \includegraphics[width=1\linewidth]{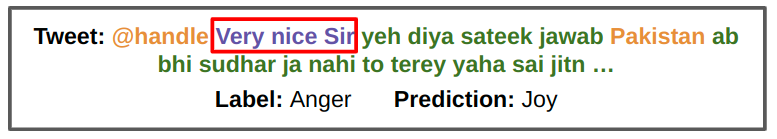}
    \caption{An example labelled \textit{anger} that was misclassified as \textit{joy} owing to the English phrase ({\color{violet}\textit{English}} - purple; {\color{teal}\textit{Hindi}} - green; {\color{orange}\textit{Other}} - orange) in the sentence having a positive connotation, even though the sentence itself conveys \textit{anger}.}
    \label{fig:countereg}
\end{figure}

We also manually look into the few instances where \textit{joy} examples were assigned a negative emotion label, and \textit{anger} examples were assigned a positive emotion label. Out of the total instances, 15 involve scenarios where either Hindi words with a negative connotation led the model to attribute a negative label to \textit{joy}, or English words with a positive connotation influenced the model to assign a positive emotion label to \textit{anger}. This suggests that examples featuring English words indicating positive emotions on their own can mislead the model into predicting a positive emotion label despite an overall negative tone in the expression (and vice versa for Hindi words), as illustrated in Figure \ref{fig:countereg}. 

On a broader scale, we examine the distribution of English, Hindi and Other tokens in the misclassified \textit{joy} and \textit{anger} examples. As seen in Table \ref{tab:norm_freq}, the normalised frequency of Hindi tokens is higher in the misclassified \textit{joy} examples than the overall distribution. Consequently, more Hindi tokens have a positive LIME score. 
Thus, \citet{mccoy-etal-2019-right}'s conclusions stated earlier are echoed here as well.
While the extreme cases where the models overgeneralise to predict an emotion label of the opposite polarity are few, there is a bias learnt in the models against predicting \textit{joy} for Hindi tokens. For examples labelled \textit{anger}, although there is less difference seen in the frequency of English tokens in the misclassified examples, more English tokens have a positive LIME score. Thus, a similar bias against predicting \textit{anger} for English could be inferred.

Overall, the fact that these associations are learnt by the models, to the extent that they can overgeneralise them, could also be seen as substantiating the sociolinguistic phenomena. If speakers tend to switch to Hindi to express negative emotions, the ability of language models to detect this reinforces the existence of such a tendency. 
This also encourages deeper engagement between sociolinguistics and interpretability, with both fields offering valuable insights to each other.

\begin{table}\centering
\captionsetup{font=small}
\small
\begin{tabular}{lrrrr}\toprule
\multicolumn{4}{c}{\textbf{Joy}} \\\cmidrule{1-4}
\multicolumn{4}{c}{Distribution of tokens in all examples} \\\cmidrule{1-4}
&\textbf{All examples} &\textbf{Correct} &\textbf{Misclassified} \\\midrule
English &0.40 &0.44 &\textbf{0.32} \\
Hindi &0.34 &0.29 &\textbf{0.44} \\
Other &0.26 &0.27 &\textbf{0.24} \\\midrule
\multicolumn{4}{c}{Distribution of tokens assigned a positive LIME score} \\\midrule
&\textbf{All examples} &\textbf{Correct} &\textbf{Misclassified} \\\midrule
English &0.43 &0.48 &\textbf{0.32} \\
Hindi &0.32 &0.28 &\textbf{0.42} \\
Other &0.25 &0.24 &\textbf{0.32} \\\midrule
\multicolumn{4}{c}{\textbf{Anger}} \\\midrule
\multicolumn{4}{c}{Distribution of tokens in all examples} \\\midrule
&\textbf{All examples} &\textbf{Correct} &\textbf{Misclassified} \\\midrule
English &0.15 &0.14 &\textbf{0.17} \\
Hindi &0.63 &0.65 &\textbf{0.61} \\
Other &0.22 &0.21 &\textbf{0.22} \\\midrule
\multicolumn{4}{c}{Distribution of tokens assigned a positive LIME score} \\\midrule
&\textbf{All examples} &\textbf{Correct} &\textbf{Misclassified} \\\midrule
English &0.15 &0.13 &\textbf{0.18} \\
Hindi &0.64 &0.68 &\textbf{0.60} \\
Other &0.21 &0.19 &\textbf{0.23} \\
\bottomrule
\end{tabular}
\caption{Normalized frequencies of \textit{English}, \textit{Hindi}, and \textit{Other} tokens for instances labeled \textit{joy} and \textit{anger} for correct and incorrect classification. Additionally, the count of tokens in each language category assigned a positive LIME score for all models.}\label{tab:norm_freq}
\end{table}

\section{Conclusion}
In this work, we use sociolinguistics theories to understand what PLMs learn when training emotion classifiers for code-mixed data. We found that the models indeed learn the differences in language use and emotional expression detailed in the sociolinguistics literature. 
Concretely, these are the associations of English tokens with positive emotions, and Hindi tokens with negative emotions. 
Adding code-mixed data to the pre-training can help augment this learning when task-specific data is scarce. However, the models can overgeneralise this learning to infrequent examples where it does not apply. In future work, it would be interesting to see if this understanding can be leveraged to help improve systems designed for code-mixed languages.

\section{Acknowledgements}
The computational resources and services used in this work were provided by the VSC (Flemish Supercomputer Center), funded by the Research Foundation - Flanders (FWO) and the Flemish Government - department EWI (for Kushal Tatariya and Miryam de Lhoneux). 
Heather Lent and Johannes Bjerva are supported by the Carlsberg Foundation, under the \textit{Semper Ardens: Accelerate} programme (project nr. CF21-0454).



\bibliography{anthology,custom}
\bibliographystyle{acl_natbib}

\appendix
\label{sec:appendix}

\section{Model Details}\label{appendix:model}

We used Huggingface to fine-tune the pre-trained language models described in Section~\ref{methods} on the emotion classification dataset. Our hyperparameters are listed in Table~\ref{tab:hyperparams} and the performance of our models over the development set are in Table~\ref{tab:model-results}, below. 

\begin{table}[!htp]\centering
\captionsetup{font=small}
\small
\begin{tabular}{lrrr}\toprule
\textbf{Hyperparameter} &\textbf{Value} \\\midrule
Dropout                 & 0.2            \\

Learning Rate           & 2e-05      \\

Number of Epochs         & 50          \\

Batch Size              & 32             \\
\bottomrule
\end{tabular}
\caption{The hyperparameters used to train all three emotion classification models. }\label{tab:hyperparams}
\end{table}


\begin{table}[!htp]\centering
\captionsetup{font=small}
\small
\begin{tabular}{lrrr}\toprule
\textbf{Model} &\textbf{Accuracy} \\\midrule
XLM-R & 0.57           \\
IndicBERT & 0.55        \\
HingRoBERTa & 0.58        \\
\bottomrule
\end{tabular}
\caption{Accuracy scores on the test sets of each pre-trained language model fine-tuned on the Hinglish emotion classification dataset.}
\label{tab:model-results}
\end{table}

\section{Label Distributions}\label{appendix:labels}

The sample of 1,000 examples used in the analysis was selected by maintaining the label distribution from the validation set. The distribution is detailed in Table \ref{tab:labels}.

\begin{table}[h]
\centering
\captionsetup{font=small}
\scriptsize
\begin{tabular}{lrrr}\toprule
\multicolumn{3}{c}{\textbf{Label Distributions}} \\\cmidrule{1-3}
\textbf{} &\textbf{Our Sample} &\textbf{Validation Set} \\\midrule
others &347 &1048 \\
joy &325 &973 \\
anger &204 &607 \\
sadness &102 &307 \\
disgust &19 &55 \\
surprise &2 &6 \\
fear &1 &4 \\\midrule
\textbf{Total} &1,000 &3,000 \\
\bottomrule
\end{tabular}
\caption{Distribution of emotion labels in our random sample versus the original validation set.}\label{tab:labels}
\end{table}

\section{Statistical Analysis}\label{appendix:anova}
\subsection{$\chi^2$}
We performed a $\chi^2$ test of independence on the samples for each model to understand the relationship between the two variables - language ID and LIME score. We constructed the contingency tables with the frequencies of how many times each language ID label - \textit{eng}, \textit{hin} and \textit{other} had a positive or a negative LIME score. We did this for the entire sample to confirm a dependency between those variables. We further examined this dependency on a more granular level by conducting the same $\chi^2$ test for examples that were predicted as \textit{joy}, \textit{anger} and \textit{sadness} by the models. The contingency table for the entire sample is in Table \ref{tab:chi-square-cont-all}, and per label is in Table \ref{tab:chisquare-cont-perlabel}.

\begin{table}[!htp]\centering
\tiny
\captionsetup{font=small}
\begin{tabular}{lrrrrrrr}\toprule
&\multicolumn{6}{c}{\textbf{Contingency Tables - All Samples}} \\\cmidrule{2-7}
&\multicolumn{2}{c}{\textbf{XLM-R}} &\multicolumn{2}{c}{\textbf{IndicBERT}} &\multicolumn{2}{c}{\textbf{HingRoBERTa}} \\\cmidrule{2-7}
&\textbf{Positive} &\textbf{Negative} &\textbf{Positive} &\textbf{Negative} &\textbf{Positive} &\textbf{Negative} \\\midrule
\textbf{English} &3658 &2264 &3840 &2082 &3728 &2194 \\
\textbf{Hindi} &5759 &4242 &5914 &4087 &6127 &3874 \\
\textbf{Other} &2709 &2203 &3281 &1631 &2843 &2069 \\
\bottomrule
\end{tabular}
\caption{$\chi^2$ contingency tables for all samples, across all models}\label{tab:chi-square-cont-all}
\end{table}

\begin{table}\centering
\tiny
\captionsetup{font=small}
\begin{tabular}{lrrrrrrr}\toprule
&\multicolumn{6}{c}{\textbf{Contingency Tables - Per Label}} \\\cmidrule{2-7}
&\multicolumn{6}{c}{\textbf{Joy}} \\\cmidrule{2-7}
&\multicolumn{2}{c}{\textbf{XLM-R}} &\multicolumn{2}{c}{\textbf{IndicBERT}} &\multicolumn{2}{c}{\textbf{HingRoBERTa}} \\\midrule
&\textbf{Positive} &\textbf{Negative} &\textbf{Positive} &\textbf{Negative} &\textbf{Positive} &\textbf{Negative} \\\midrule
\textbf{English} &1730 &759 &1669 &643 &1643 &639 \\
\textbf{Hindi} &1127 &648 &974 &500 &1061 &527 \\
\textbf{Other} &876 &668 &1010 &429 &849 &618 \\\midrule
&\multicolumn{6}{c}{\textbf{Anger}} \\\midrule
&\multicolumn{2}{c}{\textbf{XLM-R}} &\multicolumn{2}{c}{\textbf{IndicBERT}} &\multicolumn{2}{c}{\textbf{HingRoBERTa}} \\\midrule
&\textbf{Positive} &\textbf{Negative} &\textbf{Positive} &\textbf{Negative} &\textbf{Positive} &\textbf{Negative} \\\midrule
\textbf{English} &365 &173 &332 &165 &482 &288 \\
\textbf{Hindi} &1760 &689 &1487 &789 &1926 &840 \\
\textbf{Other} &473 &293 &393 &307 &551 &370 \\\midrule
&\multicolumn{6}{c}{\textbf{Sadness}} \\\midrule
&\multicolumn{2}{c}{\textbf{XLM-R}} &\multicolumn{2}{c}{\textbf{IndicBERT}} &\multicolumn{2}{c}{\textbf{HingRoBERTa}} \\\midrule
&\textbf{Positive} &\textbf{Negative} &\textbf{Positive} &\textbf{Negative} &\textbf{Positive} &\textbf{Negative} \\\midrule
\textbf{English} &248 &139 &176 &80 &233 &135 \\
\textbf{Hindi} &596 &258 &389 &187 &597 &272 \\
\textbf{Other} &187 &129 &135 &49 &169 &143 \\
\bottomrule
\end{tabular}
\caption{$\chi^2$ contingency tables for examples predicted as \textit{joy}, \textit{anger} and \textit{sadness} by each model}\label{tab:chisquare-cont-perlabel}
\end{table}

\subsection{ANOVA and Tukey HSD}

\subsubsection{Entire Sample}
\begin{table}\centering
\captionsetup{font=small}
\scriptsize
\begin{tabular}{lrr}\toprule
\multicolumn{2}{c}{\textbf{ANOVA - All Samples}} \\\cmidrule{1-2}
\textbf{Model} &\textbf{$p$-value} \\\midrule
XLM-R &2.09e-31 \\
IndicBERT &3.35e-45 \\
HingRoBERTa &3.97e-20 \\
\bottomrule
\end{tabular}
\caption{We test the null hypothesis that language ID tags and LIME scores are independent of each other using 1-Way ANOVA. This table contains the $p$-values for tests done on the entire sample.}\label{tab:anova-all-samples}
\end{table}

The $p$-values from the ANOVA results are in Table \ref{tab:anova-all-samples}. They confirm $\chi^2$ results that for the entire sample size, there is dependency between language and LIME score for all models. The key difference between our ANOVA and the $\chi^2$ tests is that, while the $\chi^2$ treats LIME score polarity as a categorical variable (positive versus negative scores), in our ANOVA we directly compute over the numerical values, ranging from -1 to 1.  

In order to better understand the relationship between languages (i.e., Hindi versus English; Hindi versus Other; English versus Other), we also performed an additional post-hoc Tukey HSD Test to test which pairs of language ID have means that are significantly different from each other. Results for all samples are in Table \ref{tab:tukey-hsd-all}. For all models, the means for Hindi and English tokens are meaningfully different from each other, and thus we can say that all models are able to distinguish between these two languages. For XLM-R, we cannot reject the null hypothesis that \textit{hin} and \textit{other} have independent distributions, and for IndicBERT, we cannot reject that \textit{eng} and \textit{other} have independent distributions. It is only for HingRoBERTa that we can reject the null hypothesis for all pairs of language ID. Thus, HingRoBERTa, having seen code-mixed data in the pre-training, is the only one that can meaningfully distinguish across \textit{eng}, \textit{hin} and \textit{other}.

\begin{table}\centering
\captionsetup{font=small}
\tiny
\begin{tabular}{lrrrrrrr}\toprule
\multicolumn{7}{c}{\textbf{Tukey HSD - All Samples}} \\\cmidrule{1-7}
\multicolumn{7}{c}{\textbf{XLMR}} \\\midrule
\textbf{group1} &\textbf{group2} &\textbf{meandiff} &\textbf{p-adj} &\textbf{lower} &\textbf{upper} &\textbf{reject} \\\midrule
en &hin &-0.013 &0 &-0.0157 &-0.0102 &\color{teal}True \\
en &other &-0.0156 &0 &-0.0188 &-0.0124 &\color{teal}True \\
hin &other &-0.0026 &0.0854 &-0.0055 &0.0003 &\color{red}False \\\midrule
\multicolumn{7}{c}{\textbf{$p$-values:} [1.218e-11, 1.218e-11, 8.538e-02]} \\\midrule
\multicolumn{7}{c}{\textbf{IndicBERT}} \\\midrule
\textbf{group1} &\textbf{group2} &\textbf{meandiff} &\textbf{p-adj} &\textbf{lower} &\textbf{upper} &\textbf{reject} \\\midrule
en &hin &-0.0144 &0 &-0.0169 &-0.0118 &\color{teal}True \\
en &other &-0.0027 &0.095 &-0.0057 &0.0003 &\color{red}False \\
hin &other &0.0117 &0 &0.009 &0.0144 &\color{teal}True \\\midrule
\multicolumn{7}{c}{\textbf{$p$-values:} [1.22E-11, 9.50E-02, 1.22E-11]} \\\midrule
\multicolumn{7}{c}{\textbf{HingRoBERTa}} \\\midrule
\textbf{group1} &\textbf{group2} &\textbf{meandiff} &\textbf{p-adj} &\textbf{lower} &\textbf{upper} &\textbf{reject} \\\midrule
en &hin &-0.0069 &0 &-0.0096 &-0.0042 &\color{teal}True \\
en &other &-0.0126 &0 &-0.0158 &-0.0095 &\color{teal}True \\
hin &other &-0.0057 &0 &-0.0086 &-0.0029 &\color{teal}True \\\midrule
\multicolumn{7}{c}{\textbf{$p$-values:} [4.29E-09, 1.22E-11, 8.12E-06]} \\
\bottomrule
\end{tabular}
\caption{Results for Tukey HSD for the entire sample size, for all models, along with the adjusted $p$-values.}\label{tab:tukey-hsd-all}
\end{table}

\subsubsection{Per Label}

\begin{table}\centering
\captionsetup{font=small}
\scriptsize
\begin{tabular}{lrrrr}\toprule
\multicolumn{4}{c}{\textbf{ANOVA - Per Label}} \\\cmidrule{1-4}
\textbf{Model} &\textbf{Joy} &\textbf{Anger} &\textbf{Sadness} \\\midrule
XLM-R &2.09e-31 &2.86e-10 &1.25e-2 \\
IndicBERT &1.53e-19 &3.20e-7 &5.57e-1 \\
HingRoBERTa &3.74e-37 &1.74e-9 &2.14e-3 \\
\bottomrule
\end{tabular}
\caption{$p$-values for 1-Way ANOVA on examples predicted as \textit{joy}, \textit{anger} and \textit{sadness} by each model.}\label{tab:anova-per-label}
\end{table}

We also conduct ANOVA tests for one positive label (\textit{joy}) and two negative labels \textit{anger, sadness}, to see whether there is agreement with the $\chi^2$ results. Table \ref{tab:anova-per-label} shows the $p$-values for each model. Both ANOVA and $\chi^2$ find dependency between language and LIME score for the \textit{joy} and \textit{anger} labels. Moreover, for \textit{sadness}, both ANOVA and $\chi^2$ also agree that there is a significant dependency of language and LIME score with HingRoBERTa, and for IndicBERT there is no dependency. Where they differ slightly is with XLM-R, where there is no dependency found with the ANOVA test, but with $\chi^2$, the $p$-value is slightly below the significance threshold.

A further fine-grained analysis of these conclusions is presented with Tukey HSD in Tables \ref{tab:tukey-hsd-joy}, \ref{tab:tukey-hsd-anger} and \ref{tab:tukey-hsd-sadness}. To summarise the results per label:
\begin{enumerate}
    \item \textbf{Joy}
    For both XLM-R and IndicBERT, \textit{hin} and \textit{other} have no meaningful difference, but do show significant distinction between \textit{hin} and \textit{eng}. HingRoBERTa, on the other hand, is able to distinguish between all language ID tags.
    \item \textbf{Anger}
    We see a significant difference between all language ID pairs and across all models for \textit{anger}.
    \item \textbf{Sadness}
    No meaningful difference is observed between \textit{hin} and \textit{eng} for both XLM-R and HingRoBERTa, and for IndicBERT, there is no meaningful difference across any of the language ID pairs. 
\end{enumerate}

\begin{table}[h]
\centering
\captionsetup{font=small}
\tiny
\begin{tabular}{lrrrrrrr}\toprule
\multicolumn{7}{c}{\textbf{Tukey HSD - Joy}} \\\cmidrule{1-7}
\multicolumn{7}{c}{\textbf{XLMR}} \\\cmidrule{1-7}
\textbf{group1} &\textbf{group2} &\textbf{meandiff} &\textbf{p-adj} &\textbf{lower} &\textbf{upper} &\textbf{reject} \\\midrule
en &hin &-0.0222 &0 &-0.0281 &-0.0164 &\color{teal}True \\
en &other &-0.0284 &0 &-0.0345 &-0.0223 &\color{teal}True \\
hin &other &-0.0062 &0.0717 &-0.0127 &0.0004 &\color{red}False \\\midrule
\multicolumn{7}{c}{\textbf{$p$-values:} [0, 0, 0.718]} \\\midrule
\multicolumn{7}{c}{\textbf{IndicBERT}} \\\midrule
\textbf{group1} &\textbf{group2} &\textbf{meandiff} &\textbf{p-adj} &\textbf{lower} &\textbf{upper} &\textbf{reject} \\\midrule
en &hin &-0.0218 &0 &-0.0277 &-0.0158 &\color{teal}True \\
en &other &-0.0169 &0 &-0.0229 &-0.011 &\color{teal}True \\
hin &other &0.0048 &0.2004 &-0.0018 &0.0114 &\color{red}False \\\midrule
\multicolumn{7}{c}{\textbf{$p$-values:} [0, 8.56E-11, 2.00E-01]} \\\midrule
\multicolumn{7}{c}{\textbf{HingRoBERTa}} \\\midrule
\textbf{group1} &\textbf{group2} &\textbf{meandiff} &\textbf{p-adj} &\textbf{lower} &\textbf{upper} &\textbf{reject} \\\midrule
en &hin &-0.0198 &0 &-0.0257 &-0.0139 &\color{teal}True \\
en &other &-0.0326 &0 &-0.0387 &-0.0266 &\color{teal}True \\
hin &other &-0.0129 &0 &-0.0194 &-0.0063 &\color{teal}True \\\midrule
\multicolumn{7}{c}{\textbf{$p$-values:} [8.54E-13, 8.42E-13, 1.15E-05]} \\
\bottomrule
\end{tabular}
\caption{Results for Tukey HSD for examples predicted as \textit{joy} by each model, along with the adjusted $p$-values.}\label{tab:tukey-hsd-joy}
\end{table}

\begin{table}\centering
\captionsetup{font=small}
\tiny
\begin{tabular}{lrrrrrrr}\toprule
\multicolumn{7}{c}{\textbf{Tukey HSD - Anger}} \\\cmidrule{1-7}
\multicolumn{7}{c}{\textbf{XLMR}} \\\cmidrule{1-7}
\textbf{group1} &\textbf{group2} &\textbf{meandiff} &\textbf{p-adj} &\textbf{lower} &\textbf{upper} &\textbf{reject} \\\midrule
en &hin &0.0076 &0.0411 &0.0002 &0.015 &\color{teal}True \\
en &other &-0.0103 &0.0152 &-0.019 &-0.0016 &\color{teal}True \\
hin &other &-0.0179 &0 &-0.0243 &-0.0115 &\color{teal}True \\\midrule
\multicolumn{7}{c}{\textbf{$p$-values:} [4.11E-02, 1.52E-02, 1.83E-10]} \\\midrule
\multicolumn{7}{c}{\textbf{IndicBERT}} \\\midrule
\textbf{group1} &\textbf{group2} &\textbf{meandiff} &\textbf{p-adj} &\textbf{lower} &\textbf{upper} &\textbf{reject} \\\midrule
en &hin &-0.0129 &0.0003 &-0.0207 &-0.0051 &\color{teal}True \\
en &other &-0.0216 &0 &-0.0308 &-0.0124 &\color{teal}True \\
hin &other &-0.0087 &0.0076 &-0.0155 &-0.0019 &\color{teal}True \\\midrule
\multicolumn{7}{c}{\textbf{$p$-values:} [3.23E-04, 1.37E-07, 1.37E-07]} \\\midrule
\multicolumn{7}{c}{\textbf{HingRoBERTa}} \\\midrule
\textbf{group1} &\textbf{group2} &\textbf{meandiff} &\textbf{p-adj} &\textbf{lower} &\textbf{upper} &\textbf{reject} \\\midrule
en &hin &0.008 &0.0184 &0.0011 &0.0149 &\color{teal}True \\
en &other &-0.0092 &0.0258 &-0.0175 &-0.0009 &\color{teal}True \\
hin &other &-0.0172 &0 &-0.0236 &-0.0107 &\color{teal}True \\\midrule
\multicolumn{7}{c}{\textbf{$p$-values:} [1.84E-02, 2.58E-02, 1.50E-09]} \\
\bottomrule
\end{tabular}
\caption{Results for Tukey HSD for examples predicted as \textit{anger} by each model, along with the adjusted $p$-values.}\label{tab:tukey-hsd-anger}
\end{table}

\begin{table}\centering
\captionsetup{font=small}
\tiny
\begin{tabular}{lrrrrrrr}\toprule
\multicolumn{7}{c}{\textbf{Tukey HSD - Sadness}} \\\cmidrule{1-7}
\multicolumn{7}{c}{\textbf{XLMR}} \\\cmidrule{1-7}
\textbf{group1} &\textbf{group2} &\textbf{meandiff} &\textbf{p-adj} &\textbf{lower} &\textbf{upper} &\textbf{reject} \\\midrule
en &hin &-0.0003 &0.9955 &-0.0092 &0.0085 &\color{red}False \\
en &other &-0.0117 &0.0323 &-0.0227 &-0.0008 &\color{teal}True \\
hin &other &-0.0114 &0.0139 &-0.0209 &-0.0019 &\color{teal}True \\\midrule
\multicolumn{7}{c}{\textbf{$p$-values:} [0.996, 0.032, 0.014]} \\\midrule
\multicolumn{7}{c}{\textbf{IndicBERT}} \\\midrule
\textbf{group1} &\textbf{group2} &\textbf{meandiff} &\textbf{p-adj} &\textbf{lower} &\textbf{upper} &\textbf{reject} \\\midrule
en &hin &-0.004 &0.5855 &-0.0135 &0.0055 &\color{red}False \\
en &other &-0.0008 &0.9868 &-0.0131 &0.0114 &\color{red}False \\
hin &other &0.0032 &0.7648 &-0.0075 &0.0139 &\color{red}False \\\midrule
\multicolumn{7}{c}{\textbf{$p$-values:} [0.585, 0.987, 0.765]} \\\midrule
\multicolumn{7}{c}{\textbf{HingRoBERTa}} \\\midrule
\textbf{group1} &\textbf{group2} &\textbf{meandiff} &\textbf{p-adj} &\textbf{lower} &\textbf{upper} &\textbf{reject} \\\midrule
en &hin &-0.0011 &0.9558 &-0.0104 &0.0081 &\color{red}False \\
en &other &-0.0149 &0.0067 &-0.0263 &-0.0034 &\color{teal}True \\
hin &other &-0.0137 &0.003 &-0.0236 &-0.0039 &\color{teal}True \\\midrule
\multicolumn{7}{c}{\textbf{$p$-values:} [0.956, 0.007, 0.003]} \\
\bottomrule
\end{tabular}
\caption{Results for Tukey HSD for examples predicted as \textit{sadness} by each model, along with the adjusted $p$-values.}\label{tab:tukey-hsd-sadness}
\end{table}

\end{document}